\documentclass[sigconf, screen, dvipsnames, prologue]{acmart}

\usepackage{amssymb}
\usepackage{graphicx}   
\usepackage{subcaption} 
\usepackage{microtype}

\usepackage{amsmath}
\usepackage{booktabs}
\usepackage{cleveref}
\usepackage{algorithm}
\usepackage{algpseudocode}
\usepackage{colortbl}
\usepackage{ulem}
\usepackage{soul}

\definecolor{blue}{rgb}{0.1, 0.3, 0.7}    
\definecolor{orange}{rgb}{1, 0.4, 0} 
\usepackage{multirow}
\usepackage{adjustbox}

\usepackage{enumitem}

\newcommand{\vvvert}{|\!|\!|}

\newtheorem{definition}{Definition}
\newtheorem{theorem}{Theorem}
\newtheorem{corollary}{Corollary}

\newtheorem{proposition}{Proposition}
\begin{document}

\title{
A Batch-Insensitive Dynamic GNN Approach to Address Temporal Discontinuity in Graph Streams
}
\author{Yang Zhou}
\affiliation{%
  \institution{Independent Researcher}
  \country{China}
}
\email{zz1999999999@outlook.com}

\author{Xiaoning Ren}
\affiliation{%
  \institution{University of Science and Technology of China}
  \city{Hefei}
  \country{China}
}
\email{hnurxn@mail.ustc.edu.cn}



\begin{abstract}
In dynamic graphs, preserving temporal continuity is critical. However, Memory-based Dynamic Graph Neural Networks (MDGNNs) trained with large batches often disrupt event sequences, leading to temporal information loss. This discontinuity not only deteriorates temporal modeling but also hinders optimization by increasing the difficulty of parameter convergence. Our theoretical study quantifies this through a Lipschitz upper bound, showing that large batch sizes enlarge the parameter search space. In response, we propose BADGNN, a novel batch-agnostic framework consisting of two core components: (1) Temporal Lipschitz Regularization (TLR) to control parameter search space expansion, and (2) Adaptive Attention Adjustment (A3) to alleviate attention distortion induced by both regularization and batching. Empirical results on three benchmark datasets show that BADGNN maintains strong performance while enabling significantly larger batch sizes and faster training compared to TGN.
Our code is available at Code: \url{https://anonymous.4open.science/r/TGN_Lipichitz-C033/} 

\end{abstract}


\keywords{Graph neural network, Temporal graph, Lipschitz bound, Attention mechanism}



\maketitle

\section{Introduction}
 
Graph representation learning has gained widespread attention due to its ability to model complex relationships in diverse applications, including social networks, recommendation systems, and bio-informatics \cite{hamilton2020graph, wu2020comprehensive,10.1145/3587268,Wu2020GraphNN,Fan2019GraphNN,Ying2018GraphCN,Hamilton2017RepresentationLO,zhou2020graph}. While early research primarily focused on static graphs \cite{kipf2016semi, hamilton2017inductive, velivckovic2017graph, kipf2016semi,Pei2020GeomGCNGG,Han2021TransformerIT,Zhang2020ResNeStSN}, real-world graph structures evolve over time, requiring Dynamic Graph Neural Networks (DGNNs) to capture these temporal dependencies \cite{kazemi2020representation, skarding2021foundations,rossi2020temporal,Manessi2017DynamicGC,Li2021DynamicGL}. DGNNs excel at capturing continuous changes in both node interactions and attributes, making them essential for modeling dynamic systems \cite{feng2024comprehensive,Skarding2020FoundationsAM,Yang2023DynamicGR}, where communication events manifest dynamically in real-time, and relationships undergo continuous evolution over time.
\begin{figure}[t]
    \centering
    \includegraphics[width=0.8\columnwidth]{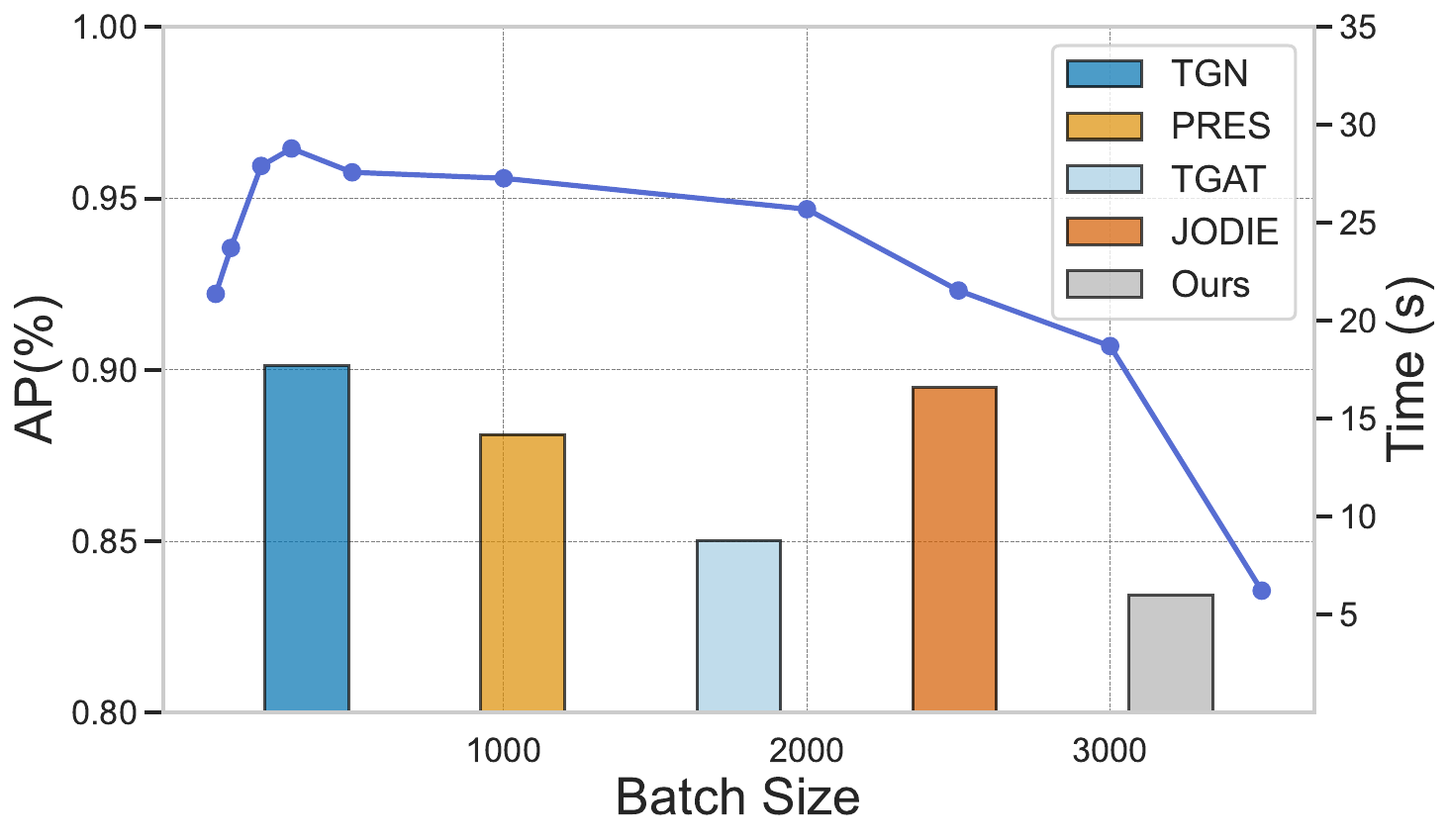}
    \caption{Line illustrates Model performance decrease with batch size increase. Bars show the time consumption of current Models and ours.}
    \label{fig:LineandBars}
\end{figure}
Despite their success, DGNNs face a fundamental challenge in batch training: temporal discontinuity. When temporal batches are processed in parallel rather than sequentially, critical temporal dependencies are disrupted (i.e., temporal discontinuity), impairing the model’s ability to capture event evolution. A common solution is to reduce batch sizes \cite{rossi2020temporal, zhou2022tgl, kumar2019predicting}, but this severely limits scalability, preventing models from handling large-scale graphs efficiently. As shown in Figure \ref{fig:LineandBars}, as the batch size increases, the model's performance significantly declines, and the computation time for existing models remains relatively large. 

To address temporal discontinuity especially on large batch, several methods have been proposed. PRES \cite{su2024pres} introduced \textbf{Memory Coherence}, which enforces \textbf{gradient consistency} within temporal batches to reduce the variance introduced by discontinuous updates. While effective, this method directly constrains gradients, which may disrupt the intrinsic structural relationships in dynamic graphs (e.g., relative influence) \cite{li2023zebra, li2024fast}. Alternative approaches include constructing batch-specific graph structures \cite{wang2024learning} and dynamically adjusting batch sizes \cite{tyagi2020taming}, but these methods either lack theoretical guarantees or introduce additional computational overhead. More importantly, these methods fail to provide a theoretical analysis and investigation of the root causes of the decrease in the performance of the model.

\begin{figure}[t]
    \centering
    \includegraphics[width=0.8\columnwidth]{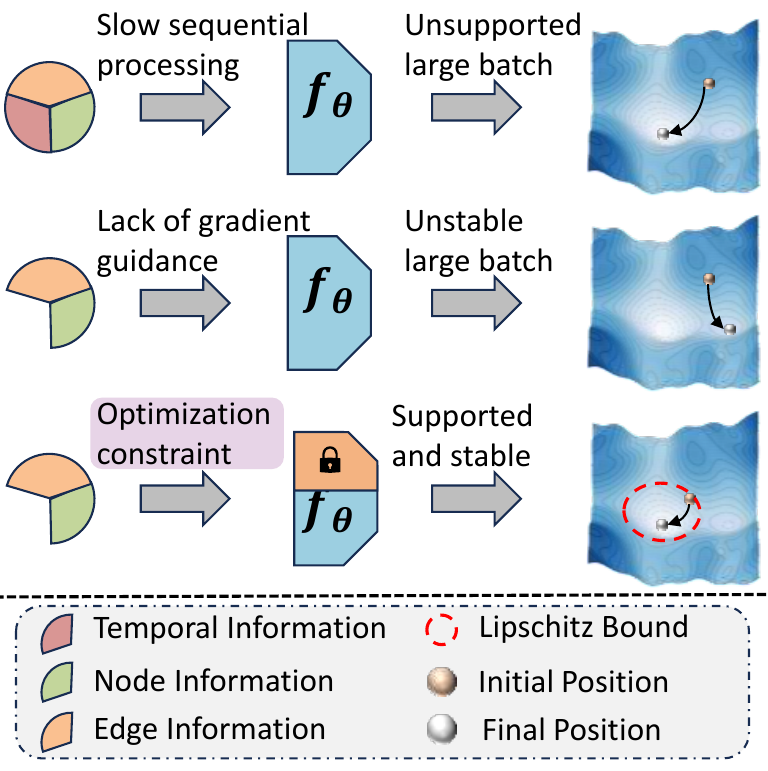}
    \caption{The top shows a model without information loss, where parameters approach an optimal solution. The middle shows a model with information loss, where parameters deviate from optimality. The bottom shows our method, which adjusts the Q–V interaction in self-attention mechanism to constrain the weight search space.
}
    \label{fig:fig_1}
\end{figure}

In this work, we propose Batch-Agnostic Dynamic Graph Neural Network (BADGNN), a robust graph representation learning method for preserving temporal dependencies on large batch. We first analyze and experimentally validate the reason why temporal discontinuity leads to performance degradation, which results from temporal information loss during batch-based training\cite{su2024pres}, hindering the model's search for the optimal position in the weight space, as shown in Figure \ref{fig:fig_1}. Additionally, we derive a theoretical Lipschitz bound, which indicates that as the batch size increases, \textbf{the model's weight search space further expands}, making it harder for the model to converge to an optimal position. The combination of the two reasons mentioned above leads to degradation of model performance. 
Motivated by our theoretical insights, we propose two methods: Temporal Lipschitz Regularization(TLR) and Adaptive Attention Adjustment(A3). TLR aims to reduce the size of the model's weight search space when batch size increases, thereby making it easier for the model to converge to the optimal position during batch-based training. A3 complements TLR by providing more fine-grained control and mitigating the side effect of overly diffuse attention caused by both large batch sizes and TLR. By using those two methods, the size of model's weight search space can be limited and thus facilitating convergence to an optimal solution while alleviating the negative impact of information loss.

 In summary, we have made the following contributions:
 \begin{enumerate}[label=\textbullet]
     \item We identify and analyze the root cause of model's performance degradation, which is stem from the addition of temporal information loss and large batch size.
     \item We derive a novel theoretical Lipschitz bound that reveals increasing the batch size leads to an expansion of the model's weight search space.
     \item We propose BADGNN to reduce the size of the model's weight search space and provide more fine-grained control.
     \item Extensive experiments on multiple real-world dynamic graph datasets show that our method consistently outperforms existing baselines under large-batch settings, achieving up to 11× larger batch sizes and a 2× speedup in training time in Wikipedia and 3.5 × larger batch sizes and a 1.4× speedup in time in Mooc compared to the original TGN.
 \end{enumerate}

\section{Related work}
\label{se:2}
\noindent
\textbf{Temporal Graph Representation Learning.}
Dynamic graph representation learning has gained significant attention due to the growing need to model evolving relationships and temporal dependencies. Early works like DynamicTriad introduced a triadic closure process to model temporal evolution in dynamic graphs, using triad structures to capture graph dynamics \cite{zhou2018dynamic}. Later advancements include dyngraph2vec, which used GNNs and RNNs to capture network dynamics by mapping nodes into vectors that reflect temporal patterns \cite{goyal2020dyngraph2vec}, and EvolveGCN, which applied recurrent neural networks like GRU and LSTM to evolve GCN parameters over time \cite{pareja2020evolvegcn}. More recent models such as JODIE focused on user-item interaction networks, employing coupled RNNs to predict future trajectories of embeddings \cite{kumar2019predicting}. APAN proposed an asynchronous propagation network for real-time temporal graph embedding, achieving scalability on large-scale graphs \cite{wang2021apan}.


Several studies have recognized the performance degradation associated with increasing batch sizes and have proposed various solutions to address this problem. \citet{wang2024learning} introduced a method that constructs graph structures within each batch, allowing the model to better capture the intrinsic relationships between samples, thus improving performance in deep representation learning. \citet{tyagi2020taming} explored the dynamic adjustment of batch sizes between different workers in distributed training environments to alleviate the performance drop caused by resource heterogeneity. Furthermore, \citet{su2024pres} focused on reducing the variance introduced by temporal discontinuity in large temporal batches by introducing Memory Coherence, a method that directly enforces gradient consistency to mitigate the effects of temporal discontinuity in dynamic graph neural networks. While this approach provides useful guidance for addressing temporal discontinuity, it has limitations. \textbf{Specifically,  direct enforcement of gradient consistency can disrupt the inherent relationships between nodes, as the weights reflect the intrinsic relationship between nodes}\cite{li2023zebra,li2024fast}. 


\textbf{Lipschitz constraint.} Pioneering work \cite{Szegedy,Zhao,Arghal,Gouk2018RegularisationON,Araujo2020FastA} derived a Lipschitz upper bound related to the model weights, or approximated the Lipschitz bound using the Jacobian matrix \cite{Jia,Qi2023UnderstandingOO,Geuchen2023UpperAL}. During the training process, a constant is artificially used to limit this Lipschitz bound. In contrast to these methods, we derive two specific terms (TLR and A3) of the Lipschitz bound and then limit the Lipschitz bound during training by incorporating TLR term into the loss function and increasing the concentration of the Softmax function's result matrix in A3 term.

\section{Preliminary}
\label{se:3}

\subsection{Problem Statement}

Dynamic graphs naturally evolve over time, as new interactions between nodes continuously emerge while older ones gradually diminish or disappear. Accurately capturing these temporal dependencies, which are often complex and fine-grained, plays a crucial role in learning graph representations that are both meaningful and robust. However, when neural networks are applied to dynamic graphs during training, a common issue known as temporal discontinuity tends to arise. This is mainly due to the manner in which events are discretely processed one at a time, which may result in inconsistencies or mismatches within the node or edge representations learned across different time points.

\subsubsection{Event-Driven Dynamic Graphs}

A dynamic graph at a given time \( t \) is represented as \( G_t = (V_t, E_t) \), where \( V_t \) denotes the set of nodes and \( E_t \) denotes the set of edges present at time \( t \). Changes in the graph’s structure are driven by events, such as the formation or deletion of edges and nodes over time. Given a sequence of such events \( B_t = \{e_1, e_2, \dots, e_n\} \), where each event \( e_i = (u_i, v_i, t_i, \delta_i) \) is associated with a specific timestamp \( t_i \), the model is expected to process all events strictly in chronological order to ensure accurate, consistent, and temporally aligned encoding of graph dynamics.

\begin{equation}
M(G_t)_{kij} = U(U(M(G_t)_k, e_i), e_j),
\end{equation}
where $M(G_t)_k$ is the model state before processing event $e_i$, and $U$ is the update function. This sequential processing ensures that the temporal relationships between events are properly preserved.

\subsubsection{The Problem of Temporal Discontinuity}

Modern dynamic graph models often rely on batch processing for efficiency, grouping multiple events together and updating the model simultaneously. And, this introduces two types of temporal discontinuity:

\textbf{Out-of-Order Updates:}

Events that need to be processed in a specific chronological order are instead applied in an incorrect sequence, resulting in improper state transitions.

\begin{equation}
M(G_t)_{kji} = U(U(M(G_t)_k, e_j), e_i).
\end{equation}

\textbf{Simultaneous Processing:}

Multiple events within a batch are processed together, ignoring their natural order:

\begin{equation}
M(G_t)_{\sigma} = U(M(G_t)_k, B_{\sigma}), \quad B_{\sigma} = \{e_i, e_j\}.
\end{equation}

This is especially common in large-batch training, where event dependencies are overlooked in favor of computational efficiency.

\subsection{Basic Notations}
For any matrix $M \in \mathbb{R}^{m \times n}$, we define $\|M\|_*$ as the largest singular value, $\|M\|_F = \left( \sum_{i,j} M_{ij}^2 \right)^{1/2}$ as the Frobenius norm, and $\|M\|_{max} = \max \left( |M_{ij}| \right)$ as the maximum element value in the matrix $M$. Moreover, for \(\mathbb{X}\) (resp. \(\mathbb{Y}\)) a vector space equipped with the norm \(\|\cdot\|_{\mathbb{X}}\) (resp. \(\|\cdot\|_{\mathbb{Y}}\)), 
the operator norm of a linear operator \(f : \mathbb{X} \to \mathbb{Y}\) will denote the quantity: $\vvvert f \vvvert_{\mathbb{X}, \mathbb{Y}} = \sup_{x \in \mathbb{X}} \frac{\|f(x)\|_{\mathbb{Y}}}{\|x\|_{\mathbb{X}}}$ and $\vvvert f \vvvert_{\mathbb{X}} = \vvvert f \vvvert_{\mathbb{X}, \mathbb{X}}$ \cite{dasoulas2021lipschitz}.

\begin{definition}
    A function \( f(x) \) is said to be Lipschitz continuous if there exists a constant \( L \) such that for any pair of points \( x \) and \( y \) in the domain of \( f(x) \), the following condition is satisfied:

\begin{equation}
    L_{\mathbb{X}, \mathbb{Y}}(f) = \sup_{x \in \mathbb{X}} \vvvert \nabla f(x) \vvvert_{\mathbb{X} \to \mathbb{Y}}\ \label{eq:Lipshitz0}.
\end{equation}
\end{definition}
Introducing a Lipschitz bound helps constrain the model's sensitivity to small changes in the input over time, preventing excessive fluctuations in the output as the graph structure evolves. 

\subsection{Attention Mechanisms}
Attention mechanisms are commonly employed in Graph Neural Networks (GNNs) due to their ability to capture global and local features. In the dynamic graph model studied here, Temporal Graph Networks (TGNs) rely heavily on attention mechanisms. The general formulation of the attention mechanism is given by:
\begin{equation}
    Att(x)=V[softmax(\frac{(QK^T)}{\sqrt{d_k}})]^T\label{eq:attn},
\end{equation}
where $softmax(M)_{ij} = \frac{e^{M_{ij}}}{\sum_{k=1}^{n} e^{M_{ik}}}$. 
Here, $X_1, X_2, X_3$ are the input matrices, and $M_q, M_k, M_v$ are the learnable parameters. The term \(d_k\) refers to the dimensionality of the key vectors.
The score function $g(x)$ is expressed as:
\begin{equation}
    g(x) = \frac{QK^T}{\sqrt{d_k}} = \frac{M_q X_1 X_2^T M_k^T}{\sqrt{d_k}}\label{eq:g(x)}.
\end{equation}
The multi-head attention mechanism is formulated:
\begin{align}\label{eq:multi_head_attn}
    MultiAtt(x)=W_0 (Att(g(x)_1) \, || \, Att(g(x)_2) \, || \, \dots \, || \, Att(g(x)_h))
\end{align}
where $||$ denotes the concatenation operation, and $W_0$ is the weight matrix applied to the concatenated outputs.

\begin{figure}
    \centering
    \includegraphics[width=0.8\columnwidth]{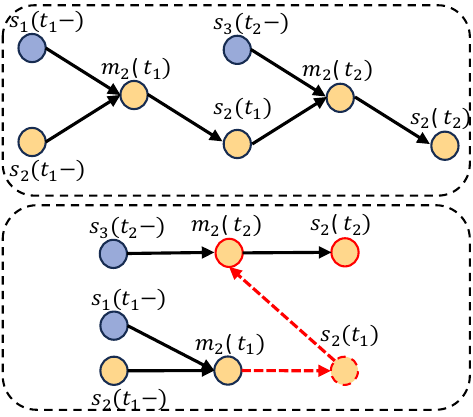}
    \caption{The top section shows the memory and states updates under a normal temporal sequence. The bottom illustrate the transition of memory states when events are sequentially processed in parallel (large batch).}
    \label{fig:loss_information}
\end{figure}

\section{Theoretical Analysis}
\label{se:Theoretical_Analysis}
\subsection{The Analysis of Temporal Discontinuity}
We first analyze the reason why temporal discontinuity leads to performance degradation. Temporal discontinuity arises mainly because the training process in batch-based manner results in \textbf{temporal information loss} (e.g., the message and state of the nodes). Define \(\nabla L_{retain}\) and \(\nabla L_{miss}\) as gradients in retaining temporal information and missing temporal information.
\begin{definition}
\label{de1}
     \textup{(\textbf{Gradient Loss})} Gradient Loss exists in batch-based process if the gradient of the missing temporal information is equal to 0 (i.e., \(\nabla L_{miss} = 0\)).
\end{definition}

\begin{proposition}
\label{proposition:temporal:loss}
Assume that the optimal solution of the model is \(w^*=[w_1, w_2]\), if there is a gradient loss, that is, \(\nabla L_{miss} = 0\), then \(w^*\) is difficult to obtain.
\end{proposition}

\textit{Simplified Proof}. According to the gradient descent algorithm, we have \(w^*=[w_1, w_2]=[w_1^{'}-\eta\nabla L_{retain}, w_2^{'}-\eta\nabla L_{miss}]\). When \(\nabla L_{miss} = 0\), it is difficult to find the optimal value for certain parameters.

Proposition \ref{proposition:temporal:loss} indicates that temporal information loss during batch-based training hinders the model’s search for the optimal position in the weight space. This reveals that the root cause of performance degradation due to temporal discontinuity is the loss of certain gradient information. 

Figure \ref{fig:loss_information} shows a scenario of information loss. When $s_1(t_1-)$ and $s_2(t_1-)$ are used to update $m_2(t_1)$, $m_2(t_2)$ is also being updated simultaneously. However, $s_2(t_1)$, which is supposed to contribute to the update of $m_w(t_2)$, does not exist at that moment. Based on Definition \ref{de1}, the gradient of information $s_2(t_1)$ ($\nabla L_{s_2(t_1)}$) is zero, which, according to Proposition \ref{proposition:temporal:loss}, leads to a suboptimal update for model parameters.

To quantify the impact of temporal discontinuity on large-batch model training, we derive the model's Lipschitz bound, which measures output sensitivity to input perturbations (e.g., misordered events). Assuming that the loss function \( l(x,y) \) is non-negative, convex, continuous, and differentiable, the overall Lipschitz constant \( L \) of the model can be expressed as the product of the Lipschitz constants of the loss function and the model components. Let \( L_F (Loss) \) and \( L_F (Model) \) denote the Lipschitz constants of the loss function and the model, respectively. Then, we have:
\begin{equation}
    L = L_F (Loss) \cdot L_F (Model).\label{eq:model_lipschitz_constant}
\end{equation}
\subsection{Lipschitz Analysis of Loss Function}
\label{se:4}
Theoretically, any non-negative, convex, continuously differentiable loss function is suitable for our analysis, and the proof of this can be found in the Appendix. But we should name one for better explanation and experiments.
In our model, we utilize the Binary Cross Entropy Loss (BCELoss) as the primary loss function for better explanation and experiments.
\noindent
BCELoss is defined as:
\begin{equation}
\begin{aligned}
    Loss =-\sum_{i=1}^n \left[ y_i \log p_i + (1 - y_i) \log (1 - p_i) \right],\label{eq:BCEloss}
\end{aligned}
\end{equation}
$x_i$ represents the model's output, $y_i$ is the ground truth label corresponding to $x_i$ and $p_i$ is the sigmoid function.

\begin{theorem}
    \label{T:BCELoss}
    The Lipschitz constant of binary cross-entropy loss $BCE_{loss}$ Eq \ref{eq:BCEloss} defined as:

\begin{equation}
L_F(Loss) = \sup_{x \in \mathbb{X}}\left| \sum_{i=1}^n (p_i - y_i) \right|.\label{eq:loss_lipschitz}
\end{equation}
\end{theorem}
The detailed proof of \textbf{Theorem \ref{T:BCELoss}} can be found in the Appendix. In Eq \ref{eq:loss_lipschitz}, $n$ denotes the batch size.
The above derivation demonstrates that the Lipschitz constant of the loss function grows with the batch size $n$, since each additional sample introduces a potential gradient term. Although the contribution of each sample is bounded ($(p_i - y_i)\in(-1,1)$), the cumulative effect leads to a looser Lipschitz bound as $n$ increases. A higher Lipschitz constant of implies that the gradient of the loss with respect to model outputs can vary more sharply, making the backpropagated gradients more volatile. As a result, the model explores a larger and potentially less stable solution space, especially when the temporal information is inconsistent across samples. This explains the degraded performance in large-batch training under temporal discontinuity.

\subsection{Constraints on Multi-head Attention}
\begin{corollary}
        \label{Multi_head}
    If each attention head is continuous and differentiable, then multi-head attention as defined in Eq \ref{eq:multi_head_attn}) is:
\begin{equation}\label{eq:model_lipschitz}
    L_F(Model) \leq \sup_{x \in \mathbb{X}}\| W_0 \|_* \sqrt{\sum_{k=1}^h \left( L_F(Att) \right)^2},
\end{equation}
\begin{align}
    L_F(Att) &\leq \sup_{x \in \mathbb{X}}\left[\frac{1}{d_k} \left[ (mn)^2 \Lambda + m^2 n^{3/2} \Lambda - \frac{2n \sigma^2}{(mn)^2}\Delta  \right]\right]^{\frac{1}{2}}\label{eq:lipschitz-bound},\\
    &\Lambda = \| M_q V^T \|_F \| M_q V^T \|_{\max} \| V \|_F \| V \|_{\max}\label{eq:IAIM},\\
    &\Delta = \left\| \left(\sqrt{S(g(x)) VV^T}\right)^T M_q V^T \right\|_*^2\label{eq:Delta},\\
   &S(g(x)) = softmax(g(x))\label{eq:softmax(g(x))}.
\end{align}

\end{corollary}
In Eq \ref{eq:lipschitz-bound}, \( m \) represents the sequence length, which corresponds to the number of neighborhood nodes considered during the aggregation process. The parameter \( n \) refers to the dimension of the input feature associated with each node. The constant \( \sigma \) is a small positive value introduced specifically to define the lower bound of the softmax function. 

The derived upper bound reveals that the Lipschitz constant of the model is influenced by multiple architectural and data-dependent terms, especially attention structures (e.g., $\Lambda$, $\Delta$). When attention scores become highly uneven or unstable,  the resulting bound becomes looser. This implies that the model's output can exhibit larger variations in response to small changes in the input. Notably, this bound highlights the critical role of attention stability in dynamic graph modeling.

\begin{proof}[Proof Sketch]

\textbf{Input.} Multi-head attention $MultiAtt(x)$ defined as:
\[
MultiAtt(x) = W_0 \left( Att(g(x)_1) \| Att(g(x)_2) \| \dots \| Att(g(x)_h) \right).
\]
\textbf{Step 1.}
Compute the derivative of $MultiAtt(x)$ with respect to $x$:
\[
\frac{\partial (MultiAtt(x))}{\partial x} = W_0 \left( \frac{\partial (Att(g(x)_1))}{\partial x} \| \dots \| \frac{\partial (Att(g(x)_h))}{\partial x} \right).
\]
\textbf{Step 2.}
 Apply the Frobenius norm inequality:
\[
\| AB \|_F^2 \leq \| A \|_*^2 \| B \|_F^2 ,
\]
\[
\left\| \frac{\partial (MultiAtt(x))}{\partial x} \right\|_F^2 \leq \| W_0 \|_*^2 \sum_{k=1}^h L_F(Att)^2.
\]
\textbf{Step 3.}
Using the Cauchy-Schwarz inequality, the H\H{o}lder's inequality and the properties of the softmax function.
\begin{align*}
&\left\| \frac{\partial (Att(x))}{\partial x_1} \right\|_F \leq \frac{1}{d_k} \left( B + C \cdot D \right),\\
&B = - \frac{2n\sigma^2}{(mn)^2} \left\| \left( \sqrt{softmax(g(x)) VV^T} \right)^T M_q V^T \right\|_*^2 ,\\
&C = \left( \sum_{i,j,p,q} (softmax(g(x))_{ij})^4 \right)^{\frac{1}{2}} 
     \left( \sum_{i,j,p,q} (M_q V^T)_{ip}^4 \right)^{\frac{1}{2}}, \\
&D = \left( \sum_{i,j,p,q} V_{jq}^4 \right)^{\frac{1}{2}} + \left( \sum_{i,j,p,q} ((softmax(g(x)) V)_{iq})^4 \right)^{\frac{1}{2}}.
\end{align*}
\textbf{Step 4.}
Applying properties of the softmax function and inequality:
\[
\sum_{i,j} M_{ij}^4  \leq (\|M\|_{\max})^2 \|M\|_F^2,
\]
\[L_F(Att) \leq \sup_{x \in \mathbb{X}}\left[\frac{1}{d_k} \left[ (mn)^2 \Lambda + m^2 n^{3/2} \Lambda - \frac{2n \sigma^2}{(mn)^2}\Delta  \right]\right]^{\frac{1}{2}}.\] 
\end{proof}
The detailed proof of \textbf{Corollary \ref{Multi_head}} can be found in the Appendix. 

\section{Method}
\label{se:5}
In this section, we propose a novel method to enhance the stability of MDGNNs during large-batch training and address the issue of temporal discontinuity. Our approach, grounded in the preceding theoretical analysis, consists of two key components: TLR and A3. By jointly optimizing these two modules, we ensure stable gradient propagation in dynamic graph settings, mitigate gradient oscillations during training, and increase batch sizes to improve training efficiency.

\subsection{Batch Sensitivity Range}

We first introduce a new metric termed the \textit{Batch Sensitivity Range (BSR)}, which quantifies the sensitivity of the model's weight search space to changes in batch size. An increase in batch size generally leads to a higher overall Lipschitz constant, causing greater fluctuations in backpropagated gradients and potentially expanding the model's weight search space explored by the model.

In the TGN model, the BSR comprises two critical terms:
\begin{equation}
\text{BSR} = \mu_1 \Lambda - \mu_2 \Delta,
\end{equation}
where \( \Lambda \) represents the interaction strength between the query matrix \( M_q \) and the value matrix \( V \), and \( \Delta \) denotes the spectral activity of the features extracted from the self-attention module—reflecting the intensity and diversity of the information the model attends to.

In MDGNNs, due to the nature of batch training, temporal dependencies in the data are inevitably compromised. This loss of sequential information increases the model’s ability to converge toward optimal solutions within an expanded model's weight search space. Therefore, controlling the BSR is especially critical for stability in batch training.

\subsection{Temporal Lipschitz Regularization}

TLR is designed to regulate the term \( \Lambda \). We introduce a regularization term into the loss function to constrain the range of \( \Lambda \):
\begin{equation}
\mathcal{L} = \mathcal{L}_{\text{regular}} + \lambda_{\text{TLR}} R(X).
\end{equation}

However, directly incorporating the entire \( \Lambda \) as a regularization term is impractical in the TGN setting. Many terms are redundant, and the full \( \Lambda \) leads to excessively large values, which may dominate the loss and make it difficult to tune \( \lambda_{\text{TLR}} \). Additionally, terms such as \( \| M_q V^T \|_{\max} \) and \( \| V \|_{\max} \) involve max operations that hinder backpropagation. 

Thus, we adopt a simplified yet effective approximation to serve as the regularization term. This form retains the essential constraints on attention distribution while significantly reducing computational complexity:
\begin{equation}
\label{eq:regularizationterm}
R(X) = \| M_q V^T \|_F \cdot \| V \|_F.
\end{equation}
The regularization term imposes a directional constraint between the current node weight representation $M_q$ and its neighbor representation $V$, encouraging them to be approximately orthogonal in the feature space. This design effectively limits the set of representational directions the model can adopt, thereby constraining the admissible parameter configurations. In addition, by enforcing such geometric structure, the regularization term implicitly reduces the volume of the parameter search space. This constraint introduces a structural bias that steers the model toward more stable update trajectories in high-dimensional space, mitigating overfitting and parameter drift commonly encountered in large-batch training, and ultimately enhancing stability.

\subsection{Adaptive Attention Adjustment}

While the simplified and direct approach of using \( \Lambda \) as a regularization term improves computational efficiency, it may lead to potential issues due to its oversimplification. Given the complexity and variability of model training, a more flexible control mechanism is required. In our framework, A3 dynamically adjusts the attention weights to mitigate the negative effects of the regularization, leveraging \( \Delta \) to counterbalance its impact. From Eq. \ref{eq:lipschitz-bound}, it can be observed that \( \Lambda \) and \( \Delta \) share many common components but with opposite coefficients, and that \( \mu_2 \ll \mu_1 \). Hence, we can achieve partial cancellation of the regularization effect by manipulating different terms in the softmax function \( g(x) \). Direct modification of softmax inputs is impractical. Therefore, we indirectly enhance the singular values of the softmax matrix. We redefine the score function used in the attention mechanism as:
\begin{equation}
\label{eq:Attentionterm}
g(x) = mn \cdot \frac{QK^T}{\sqrt{d_k}}.
\end{equation}
By applying \( d_k = \frac{d_k}{(mn)^2} \), where \( d_k \) is the dimensionality of the key vectors in the attention mechanism, \( m \) is the sequence length (the number of sampled neighbors in TGN), and \( n \) is the input feature dimension. A smaller \( d_k \) results in a larger \( g(x) \), which, after normalization through the \( S(g(x)) \) function, yields more concentrated attention weights. Dividing \( d_k \) increases the magnitude of \( g(x) \), leading to more concentrated attention weights after applying the softmax function. This adjustment counteracts the dispersion of attention caused by large batch sizes and regularization, thereby enhancing the model's focus on relevant features. 

Furthermore, A3 introduces a non-uniform scaling transformation in the attention space. This transformation dynamically applies varying degrees of stretching or compression to the key representations along different directions, reshaping the originally isotropic attention space—typically a high-dimensional structure—into a time-dependent ellipsoid or a more complex asymmetric shape. Consequently, the similarity measure in attention computation is no longer solely determined by the standard inner product; instead, it incorporates weighted amplification or suppression along specific directions, effectively performing a controllable directional projection adjustment in the space. This mechanism significantly limits the effective region of the attention distribution, allowing the model to focus more on specific geometric regions during inference, thereby further compressing the search space and enhancing the model's expressive efficiency in dynamic environments.

\begin{table}[h]
\centering
\caption{The statistics of datasets, where $n$ represents the number of nodes, $m$ denotes the number of edges, $|T|$ indicates the number of timestamps or snapshots, and $d_e$ refers to the dimension of edge features.}
\label{tab:data_datil}

\begin{tabular}{c|c|c|c|c|c}
\hline
Datasets  & Domain        & $n$     & $m$      & $|T|$    & $d_e$ \\ \hline
Wikipedia & Social       & 9227  & 157474 & 152757 & 172  \\ \hline
Reddit    & Social       & 10984 & 672447 & 669065 & 172  \\ \hline
Mooc      & Interaction  & 7144  & 411749 & 345600 & 4    \\ \hline
\end{tabular}%

\end{table}

\section{Experimental Results}
\subsection{Experimental Setup}
\label{se:6}
\subsubsection{Datasets}
To evaluate the performance of the proposed method, we conducted experiments on three publicly available real-world datasets: Wikipedia, Reddit, and MOOC\cite{kumar2019predicting}, each provided under its respective license. Details of those datasets can be found in (Table \ref{tab:data_datil}). These datasets were selected due to their dynamic nature and diverse interaction patterns, providing a rigorous testbed for assessing temporal continuity in dynamic graph models. 

\subsubsection{Baseline Models}

In order to evaluate the effectiveness of our proposed method, we compare it against the following popular baseline models:

\begin{enumerate}[label=\textbullet]

\item \textbf{Temporal Graph Networks (TGN)} \cite{rossi2020temporal} are designed to efficiently learn from dynamic graphs by modeling the temporal nature of interactions between nodes. TGN incorporates memory modules and graph-based operators, allowing it to capture long-term temporal dependencies and outperform traditional static graph models. The framework is both efficient and scalable, providing a robust solution for dynamic graph learning tasks.

\item \textbf{Temporal Graph Attention Networks (TGAT)} \cite{xu2020inductive} use a self-attention mechanism to aggregate features from temporal and topological neighborhoods. By utilizing time-encoded embeddings, TGAT is able to capture the evolving nature of node embeddings over time. The model is capable of inductively inferring embeddings for new nodes and works effectively for tasks such as node classification and link prediction in temporal graphs.

\begin{figure}[t]
  \centering
  \includegraphics[width=0.45\textwidth]{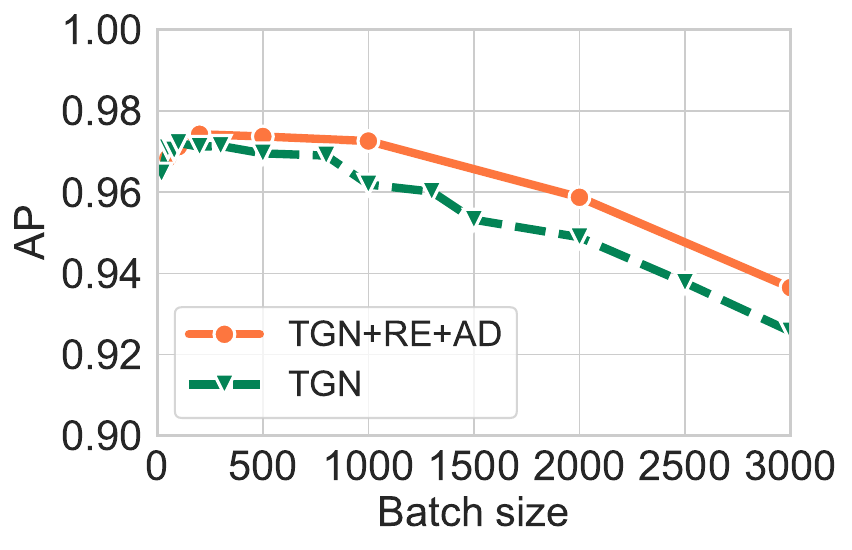}
  \caption{This figure illustrates the AP of TGN and our proposed method, BADGNN, evaluated under batch sizes ranging from 50 to 3500.}
  \label{fig:wikipedia1_performance}
\end{figure}

\item \textbf{Predictive Regularization Strategy (PRES)} \cite{su2024pres} aims to address the issue of temporal discontinuity in Memory-based Dynamic Graph Neural Networks (MDGNNs). PRES incorporates a memory coherence learning objective and an iterative prediction-correction scheme, enabling MDGNNs to process larger temporal batches without compromising generalization performance. This method improves training efficiency, achieving a significant speed-up in large-scale dynamic graph tasks.

\end{enumerate}

\subsubsection{Implementation Details}

The selected models for our experiments include TGN, TGAT, and TGN with the PRES module, as well as three variations: TGN-TLR, which incorporates only the regularization term; TGN-A3, which applies only the attention-adjustment; and BADGNN, which combines both the regularization term and attention-adjustment. All experiments were conducted in an environment equipped with an NVIDIA 3080ti GPU, running Python 3.8, CUDA version 11.1, and PyTorch version 1.8.1.  All baseline models (TGN, PRES, TGAT) adopt the same optimizer (Adam), learning rate (0.0005/3e-4 for transductive/inductive tasks), and training epochs. No learning rate scheduling is applied. In addition, a dropout rate of 0.1 was applied in all experiments. The number of neighboring nodes was fixed at 1, and the network architecture consisted of 1 layers in all experiments to ensure consistency across different configurations. Other Settings are the same as those of TGN \cite{rossi2020temporal}.

\subsubsection{Evaluation Metrics}
We used two main evaluation metrics to assess the performance of the model: Average Precision (AP) and Area Under the Curve (AUC). Both metrics help evaluate the model's ability to make accurate predictions. For training efficiency, we also measured the time taken to train the model for one epoch. AP and AUC: Higher values indicate better model performance. Time Consumption: Lesser training time indicates higher efficiency.

\subsection{Overall Performance}
Our proposed model, BADGNN, demonstrates superior performance across various dynamic graph datasets in terms of both predictive accuracy and computational efficiency. It consistently achieves high Average Precision (AP) while accelerating training, making it suitable for practical, large-scale scenarios.

\begin{table}[t]
\centering
\caption{AP (\%), Batch Size and Time consumption (50 epoch) for different models across various datasets. (The percentages in parentheses represent the comparison between our method and TGN.)}
    \label{tab:performance_compare}
\setlength{\tabcolsep}{6pt}
\begin{tabular}{ll|c|c}
\hline
\multicolumn{2}{c|}{Model}                                & Wikipedia               & Mooc                   \\ \hline
\multicolumn{1}{c|}{\multirow{2}{*}{AP(\%)}}     & TGN    & 96.4                    & 72.9                   \\ \cline{2-4} 
\multicolumn{1}{c|}{}                            & BADGNN & \textbf{96.2(-0.2\%)}   & \textbf{73.2(+0.4\%)} \\ \hline
\multicolumn{1}{c|}{\multirow{2}{*}{Batch Size}} & TGN    & 300                     & 1000                   \\ \cline{2-4} 
\multicolumn{1}{c|}{}                            & BADGNN & \textbf{3500(+1066.7\%)} & \textbf{3500(+250\%)}   \\ \hline
\multicolumn{1}{c|}{\multirow{2}{*}{Time(s)}}    & TGN    & 641                     & 1845                   \\ \cline{2-4} 
\multicolumn{1}{c|}{}                            & BADGNN & \textbf{297(-53.7\%)}    & \textbf{1291(-30\%)}    \\ \hline
\end{tabular}%
\end{table}

\begin{table}[t]
    \centering
    \caption{AP (\%) and AUC(\%) for different models across various datasets at a batch size of 3500. In each row, the best result is highlighted in bold, and the runner-up is underlined.}
    \label{tab:performance_speedup}
    \setlength{\tabcolsep}{2pt} 
    \begin{tabular}{l|cc|cc|cc}
        \toprule
        \textbf{Dataset} & \multicolumn{2}{c|}{\textbf{REDDIT}} & \multicolumn{2}{c|}{\textbf{WIKI}} & \multicolumn{2}{c}{\textbf{MOOC}} \\
        \hline
        \textbf{Model} & \textbf{AP(\%)} & \textbf{AUC(\%)} & \textbf{AP(\%)} & \textbf{AUC(\%)} & \textbf{AP(\%)} & \textbf{AUC(\%)} \\
        \hline
        TGN         & \underline{98.2} &98.2  & 83.6$\pm$0.1 & 86.4$\pm$0.1 & 67.5$\pm$0.1 & 69.8$\pm$0.1\\
        TGAT   & 93.3$\pm$0.1 & 92.1 & 72.1  & 70.3 & - & -\\
        PRES      & 91.8 & 91.8 & 89.7$\pm$0.1 & 89.7 & 65.3 & 67.6 \\
        TGN-TLR & 98.3 & 98.2  & 95.6 & 95.3 & \textbf{74.0$\pm$0.1} & \textbf{77.3$\pm$0.1} \\
        TGN-A3       & 98.3 &\underline{98.2} & \underline{96.2} & \underline{96.0$\pm$0}.0 & 72.3 &76.0$\pm$0.1 \\
        BADGNN  & \textbf{98.4} & \textbf{98.3} & \textbf{96.2} & \textbf{96.0} & \underline{73.2$\pm$0.1} & \underline{76.6$\pm$0.1} \\
        \bottomrule
    \end{tabular}%
\end{table}

\begin{table}[t]
\centering
\caption{Transductive Performance on Reddit, Wikipedia, and MOOC datasets. ROC-AUC(\%).}
  \label{tab:performance}
\setlength{\tabcolsep}{12pt} 
\begin{tabular}{lccc}
\hline
\multicolumn{1}{l}{Model} & Reddit            & Wikipedia     & MOOC          \\ \hline
TGN                       & 60.8              & 86.3          & 62.3$\pm$2.6      \\
TGAT                      & 59.9              & 80.8          & -             \\
PRES                      & \textbf{65.0$\pm$1.2} & 84.7$\pm$1.8      & 56.9$\pm$3.3      \\
TGN-TLR                   & 60.0              & 87.3          & 61.6$\pm$0.1      \\
TGN-A3                    & 60.1$\pm$0.1          & 89.7          & 63.6          \\
BADGNN                   & 62.3$\pm$0.1          & \textbf{91.2} & \textbf{63.7} \\ \hline
\end{tabular}%
\end{table}

\textbf{BADGNN Performance with Large Batch Sizes.}
A notable strength of BADGNN lies in its ability to handle large batch sizes while maintaining high performance. As shown in Table \ref{tab:performance_compare}, BADGNN successfully maintains performance at batch sizes up to 3500—more than 11× and 3.5× larger than those used by TGN on Wikipedia and Mooc respectively—demonstrating strong robustness to batch size scaling. In Table~\ref{tab:performance_speedup}, where all models are evaluated under the same large batch size of 3500, our proposed methods consistently outperform other models in terms of both AP and AUC across most cases. Notably, we observe that on the Reddit dataset, most models still achieve relatively high performance. This may be attributed to the fact that Reddit exhibits a much longer temporal span compared to the other datasets, which makes it less sensitive to temporal discontinuities caused by large batch training. As a result, the negative impact of increasing the batch size is less pronounced on Reddit.

\textbf{Stability and Comparison with Other Models.}
BADGNN exhibits a clear and consistent advantage over TGN and PRES, especially as the batch size increases, where other models suffer noticeable performance drops. BADGNN consistently outperforms both TGN and PRES across a wide range of batch sizes, as shown in Table \ref{tab:performance_batch}. This advantage becomes particularly significant under large-batch training settings. Figure \ref{fig:wikipedia1_performance} further reinforces this by directly comparing TGN with and without our proposed modifications, illustrating that BADGNN effectively mitigates the negative effects of increasing batch sizes. This clearly demonstrates that BADGNN is not only more computationally efficient but also more robust and stable than existing models in handling large-scale dynamic graph data.

\textbf{Transductive Performance Comparison.}
Table~\ref{tab:performance} reports the transductive performance of our proposed models (TGN-TLR, TGN-A3, and BADGNN) compared to baseline models (TGN, TGAT, and PRES) across three dynamic graph datasets. As shown, BADGNN achieves the best overall performance on both the Wikipedia and MOOC datasets, with an AUC of 91.2\% and 63.7\%, respectively. Notably, BADGNN surpasses the original TGN model by a significant margin, especially on Wikipedia where the improvement reaches nearly 5\%. On the Reddit dataset, while PRES achieves the highest AUC, BADGNN still outperforms TGN and TGAT, indicating its stable transductive performance.

The experimental results align with the theoretical analysis of BADGNN. By enhancing the regularization term and the attention mechanism, BADGNN maintains training stability and high efficiency even with increased batch sizes. The mechanisms proposed in the theoretical framework are validated through experiments, particularly showing that BADGNN mitigates the adverse effects of larger batch sizes. Overall, the experimental and theoretical results demonstrate that BADGNN not only adapts well to large-batch training but also maintains superior performance on large-scale dynamic graph datasets.

\subsection{Ablation Study}
In this section, we conduct ablation studies to assess the impact of key components of our proposed method.
\subsubsection{Ablation Study on Regularization Term}

\begin{figure*}[ht]
  \centering
  \begin{subfigure}{0.45\linewidth}
    \includegraphics[width=\linewidth]{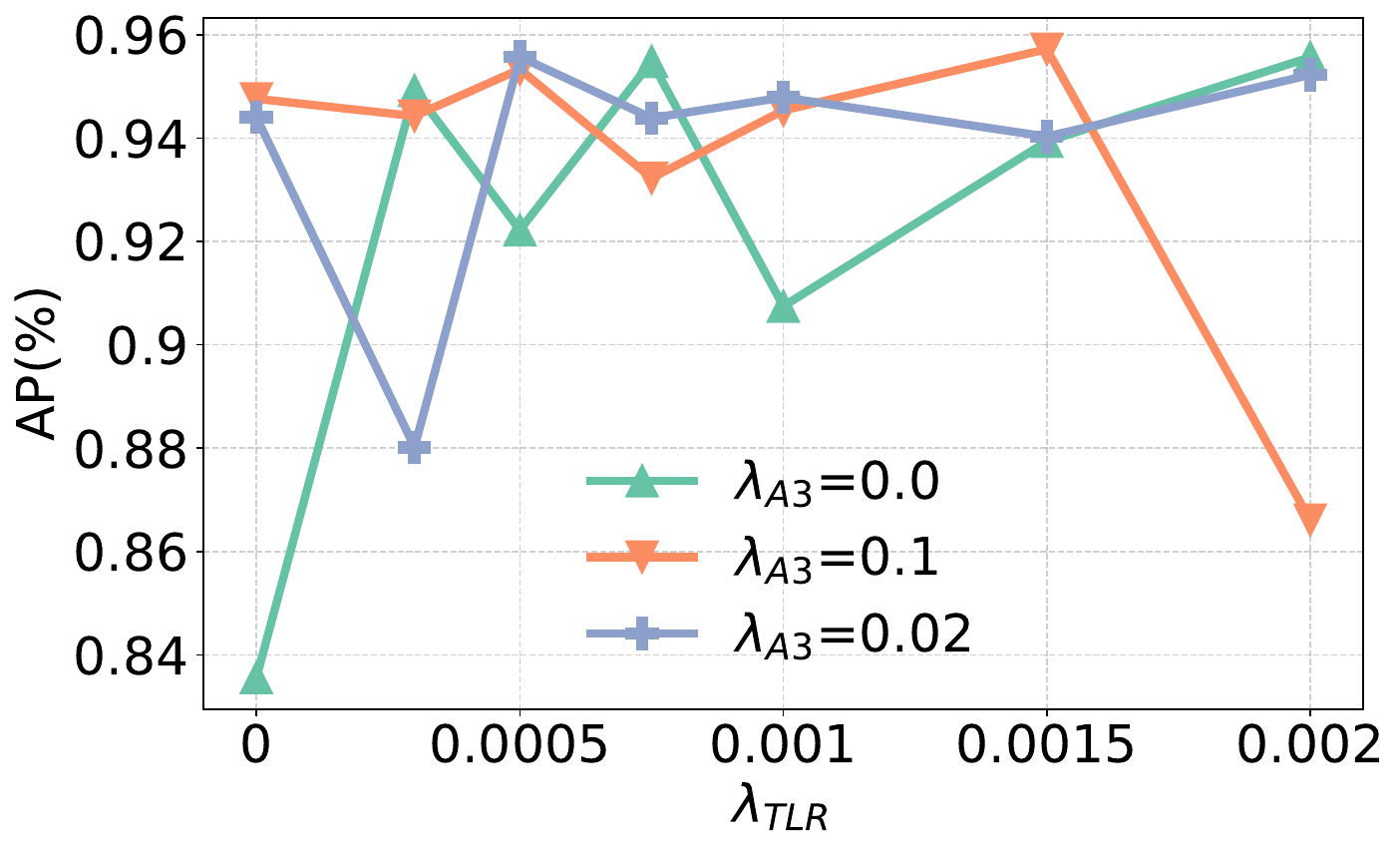} 
    \caption{AP (\%) with respect to TLR while the coefficient of A3 remains stable.}
    \label{fig:short-a}
  \end{subfigure}
  \hfill
  \begin{subfigure}{0.45\linewidth}
    \includegraphics[width=\linewidth]{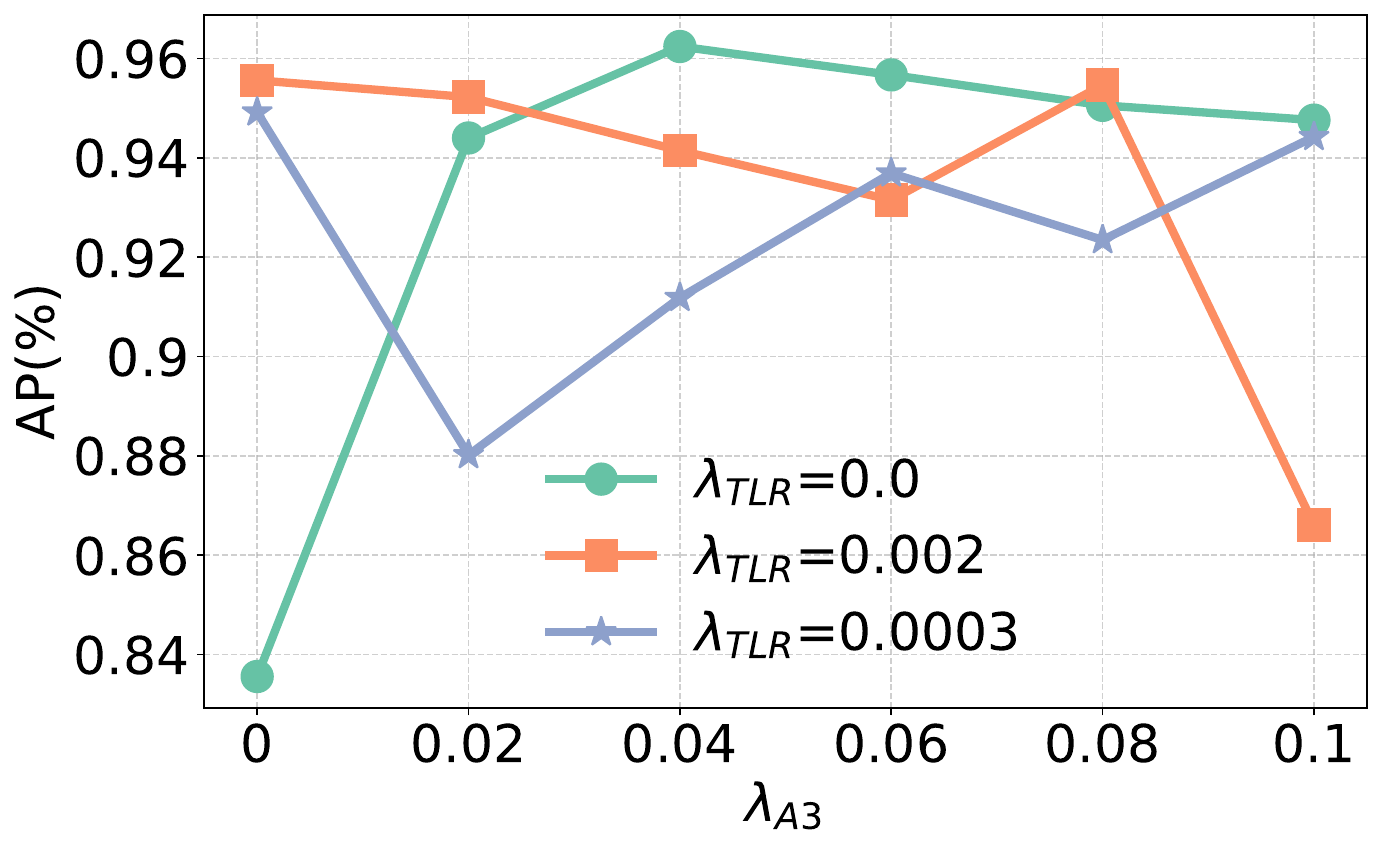} 
    \caption{AP (\%) with respect to A3 while the coefficient of TLR remains stable.}
    \label{fig:short-b}
  \end{subfigure}
  \caption{Analysis of Model Performance Sensitivity to coefficient of A3 and TLR.}
  \label{fig:short}
\end{figure*}


\begin{table}[t]
\centering
\caption{Comparative performance (AP\%) analysis of BADGNN, PRES, and TGN across varying batch sizes on the MOOC and WIKI datasets.}
\label{tab:performance_batch}

\setlength{\tabcolsep}{4pt} 
\begin{tabular}{c|ccc|ccc}
\hline
\textbf{Dataset}    & \multicolumn{3}{c|}{\textbf{MOOC}}             & \multicolumn{3}{c}{\textbf{WIKI}}              \\ \hline
\textbf{Batch} & \textbf{BADGNN} & \textbf{PRES} & \textbf{TGN} & \textbf{BADGNN} & \textbf{PRES} & \textbf{TGN} \\ \hline
100                 & 82.9            & 66.3          & 80.6         & 96.5            & 91.9          & 95.2         \\
500                 & 80.7            & 65.2          & 79.2         & 96.8            & 89.3          & 96.1         \\
1000                & 76.5            & 61.6          & 72.9         & 97.0            & 90.3          & 95.8         \\
2000                & 75.5            & 63.4          & 72.0         & 96.9            & 87.8          & 96.0         \\
2500                & 74.2            & 64.1          & 72.6         & 96.5            & 88.8          & 94.7         \\
3000                & 74.3            & 59.8          & 68.7         & 96.0            & 80.1          & 94.5         \\
3500                & 73.2            & 65.3          & 67.5         & 96.2            & 89.7          & 83.6         \\ \hline
\end{tabular}
\end{table}

We conducted an ablation study to examine the contribution of the proposed regularization term by removing it from our BADGNN model, resulting in a variant that only employs attention adjustment during training (denoted as TGN-A3). We evaluated this variant across all three datasets—Wikipedia, Reddit, and MOOC—and reported the AP and AUC metrics in Table \ref{tab:performance_speedup} and Table \ref{tab:performance}. The results clearly demonstrate that the absence of the regularization term leads to a consistent decline in performance across datasets, with the most pronounced gap observed on the Reddit dataset under large batch sizes. This suggests that the regularization term plays a crucial role in enhancing the model’s temporal robustness and mitigating overfitting to instantaneous interactions. 

\subsubsection{Ablation Study on Attention Adjustment}
To assess the effectiveness of the proposed attention adjustment mechanism, we removed it from the BADGNN model while keeping the regularization term, resulting in a simplified version referred to as TGN-TLR. Performance comparisons across the three datasets are shown in Table \ref{tab:performance_speedup} and Table \ref{tab:performance}. The results indicate that the removal of the attention adjustment leads to a noticeable degradation in both AP and AUC. This performance drop underscores the importance of attention adjustment in adapting to evolving graph structures and maintaining model discriminability. 

We conduct a series of experiments to evaluate how different configurations of two key hyperparameters—$\lambda_{TLR}$, the influence coefficient of the regularization term, and $\lambda_{A3}$, the influence coefficient of the attention adjustment—affect the model’s performance. All experiments are carried out on the Wikipedia dataset with a fixed batch size of 3500. The results are presented in Figure~\ref{fig:short}.

\subsection{Hyperparameter Sensitivity Analysis}
To evaluate the robustness of BADGNN with respect to its key hyperparameters, we conduct a sensitivity analysis on the TLR weight ($\lambda_{TLR}$) and the coefficient of A3 ($\lambda_{A3}$). The results are visualized in Figure \ref{fig:short-a} and Figure \ref{fig:short-b}, where we show the AP under varying hyperparameter settings.

In Figure \ref{fig:short-a}, we fix ($\lambda_{A3}$) at three representative values (0, 0.1, and 0.02), and observe the model’s performance as $\lambda_{TLR}$ increases from 0 to 0.002. When $\lambda_{A3}$ is non-zero, introducing a moderate value of $\lambda_{TLR}$ significantly improves the performance. For instance, when $\lambda_{A3}=0.02$, AP increases from 0.94 (when $\lambda_{TLR}=0$) to a peak of approximate 0.96 at $\lambda_{TLR}=0.0005$. However, like the analysis in Section \ref{se:5}, excessively large $\lambda_{TLR}$ may lead to performance drops. Notably, when $\lambda_{A3}=0$ (no attention adjustment), TLR leads to substantial gains, suggesting that the regularization term alone can already bring noticeable improvements.

In Figure \ref{fig:short-b}, we fix $\lambda_{TLR}$ at 0, 0.002, and 0.0003, and vary $\lambda_{A3}$ from 0 to 0.1. The results indicate that performance generally improves with increasing $\lambda_{A3}$ up to a point. For example, when $\lambda_{TLR}=0$, AP peaks at slightly above 0.96 when $\lambda_{A3}=0.04$. However, further increases in $\lambda_{A3}$ may hurt performance, likely due to overly aggressive modification of attention weights.

Overall, BADGNN performs consistently well across a wide range of $\lambda_{TLR}$ and $\lambda_{A3}$ values, showing strong robustness. The best performance is observed when both hyperparameters take moderate values, such as $\lambda_{A3} \approx 0.04$ and $\lambda_{TLR} \approx 0.0005$, where the model achieves a balanced synergy between regularization control and attention adaptation.
\section{Conclusion}
\label{se:7}
In this paper, we investigate the impact of large-batch training on DGNNs, particularly how it affects the model’s representation stability and temporal expressiveness. While large batches improve training efficiency, they also introduce optimization challenges and distort attention dynamics, leading to degraded performance. To address these issues, we propose two complementary techniques: TLR, which geometrically constrains the feature space by encouraging orthogonality between node representations, thereby reducing the parameter search space and stabilizing optimization; and A3, that adaptively reshapes the attention distribution through a non-uniform scaling of key representations. This adjustment helps the model maintain performance even under large-batch conditions, reducing the negative impact of batch size on representation quality. Together, these methods enhance robustness and temporal representation learning on dynamic graphs, improving model effectiveness without requiring explicit event sequence modeling or reordering. As a result, they enable more stable and efficient training in dynamic graph scenarios where large batches are commonly used.

\bibliographystyle{ACM-Reference-Format}
\bibliography{sample-base}


\begin{thebibliography}{43}


\ifx \showCODEN    \undefined \def \showCODEN     #1{\unskip}     \fi
\ifx \showISBNx    \undefined \def \showISBNx     #1{\unskip}     \fi
\ifx \showISBNxiii \undefined \def \showISBNxiii  #1{\unskip}     \fi
\ifx \showISSN     \undefined \def \showISSN      #1{\unskip}     \fi
\ifx \showLCCN     \undefined \def \showLCCN      #1{\unskip}     \fi
\ifx \shownote     \undefined \def \shownote      #1{#1}          \fi
\ifx \showarticletitle \undefined \def \showarticletitle #1{#1}   \fi
\ifx \showURL      \undefined \def \showURL       {\relax}        \fi
\providecommand\bibfield[2]{#2}
\providecommand\bibinfo[2]{#2}
\providecommand\natexlab[1]{#1}
\providecommand\showeprint[2][]{arXiv:#2}

\bibitem[Araujo et~al\mbox{.}(2020)]%
        {Araujo2020FastA}
\bibfield{author}{\bibinfo{person}{Alexandre Araujo}, \bibinfo{person}{Benjamin N{\'e}grevergne}, \bibinfo{person}{Yann Chevaleyre}, {and} \bibinfo{person}{Jamal Atif}.} \bibinfo{year}{2020}\natexlab{}.
\newblock \showarticletitle{Fast \& Accurate Method for Bounding the Singular Values of Convolutional Layers with Application to Lipschitz Regularization}.
\newblock \bibinfo{journal}{\emph{ArXiv}}  \bibinfo{volume}{abs/2006.08391} (\bibinfo{year}{2020}).
\newblock
\urldef\tempurl%
\url{https://api.semanticscholar.org/CorpusID:219686989}
\showURL{%
\tempurl}


\bibitem[Arghal et~al\mbox{.}(2022)]%
        {Arghal}
\bibfield{author}{\bibinfo{person}{R. Arghal}, \bibinfo{person}{E. Lei}, {and} \bibinfo{person}{S.~S. Bidokhti}.} \bibinfo{year}{2022}\natexlab{}.
\newblock \showarticletitle{Robust graph neural networks via probabilistic Lipschitz constraints}. In \bibinfo{booktitle}{\emph{Learning for Dynamics and Control Conference}}. \bibinfo{pages}{1073–1085}.
\newblock


\bibitem[Dasoulas et~al\mbox{.}(2021)]%
        {dasoulas2021lipschitz}
\bibfield{author}{\bibinfo{person}{George Dasoulas}, \bibinfo{person}{Kevin Scaman}, {and} \bibinfo{person}{Aladin Virmaux}.} \bibinfo{year}{2021}\natexlab{}.
\newblock \showarticletitle{Lipschitz normalization for self-attention layers with application to graph neural networks}. In \bibinfo{booktitle}{\emph{International Conference on Machine Learning}}. PMLR, \bibinfo{pages}{2456--2466}.
\newblock


\bibitem[Fan et~al\mbox{.}(2019)]%
        {Fan2019GraphNN}
\bibfield{author}{\bibinfo{person}{Wenqi Fan}, \bibinfo{person}{Yao Ma}, \bibinfo{person}{Qing Li}, \bibinfo{person}{Yuan He}, \bibinfo{person}{Yihong~Eric Zhao}, \bibinfo{person}{Jiliang Tang}, {and} \bibinfo{person}{Dawei Yin}.} \bibinfo{year}{2019}\natexlab{}.
\newblock \showarticletitle{Graph Neural Networks for Social Recommendation}.
\newblock \bibinfo{journal}{\emph{The World Wide Web Conference}} (\bibinfo{year}{2019}).
\newblock
\urldef\tempurl%
\url{https://api.semanticscholar.org/CorpusID:67769538}
\showURL{%
\tempurl}


\bibitem[Feng et~al\mbox{.}(2024)]%
        {feng2024comprehensive}
\bibfield{author}{\bibinfo{person}{ZhengZhao Feng}, \bibinfo{person}{Rui Wang}, \bibinfo{person}{TianXing Wang}, \bibinfo{person}{Mingli Song}, \bibinfo{person}{Sai Wu}, {and} \bibinfo{person}{Shuibing He}.} \bibinfo{year}{2024}\natexlab{}.
\newblock \showarticletitle{A Comprehensive Survey of Dynamic Graph Neural Networks: Models, Frameworks, Benchmarks, Experiments and Challenges}.
\newblock \bibinfo{journal}{\emph{arXiv preprint arXiv:2405.00476}} (\bibinfo{year}{2024}).
\newblock


\bibitem[Geuchen et~al\mbox{.}(2023)]%
        {Geuchen2023UpperAL}
\bibfield{author}{\bibinfo{person}{Paul Geuchen}, \bibinfo{person}{Thomas Heindl}, \bibinfo{person}{Dominik St{\"o}ger}, {and} \bibinfo{person}{Felix Voigtlaender}.} \bibinfo{year}{2023}\natexlab{}.
\newblock \showarticletitle{Upper and lower bounds for the Lipschitz constant of random neural networks}.
\newblock \bibinfo{journal}{\emph{ArXiv}}  \bibinfo{volume}{abs/2311.01356} (\bibinfo{year}{2023}).
\newblock
\urldef\tempurl%
\url{https://api.semanticscholar.org/CorpusID:264935036}
\showURL{%
\tempurl}


\bibitem[Gouk et~al\mbox{.}(2018)]%
        {Gouk2018RegularisationON}
\bibfield{author}{\bibinfo{person}{Henry Gouk}, \bibinfo{person}{Eibe Frank}, \bibinfo{person}{Bernhard Pfahringer}, {and} \bibinfo{person}{Michael~J. Cree}.} \bibinfo{year}{2018}\natexlab{}.
\newblock \showarticletitle{Regularisation of neural networks by enforcing Lipschitz continuity}.
\newblock \bibinfo{journal}{\emph{Machine Learning}}  \bibinfo{volume}{110} (\bibinfo{year}{2018}), \bibinfo{pages}{393 -- 416}.
\newblock
\urldef\tempurl%
\url{https://api.semanticscholar.org/CorpusID:4811672}
\showURL{%
\tempurl}


\bibitem[Goyal et~al\mbox{.}(2020)]%
        {goyal2020dyngraph2vec}
\bibfield{author}{\bibinfo{person}{Palash Goyal}, \bibinfo{person}{Sujit~Rokka Chhetri}, {and} \bibinfo{person}{Arquimedes Canedo}.} \bibinfo{year}{2020}\natexlab{}.
\newblock \showarticletitle{dyngraph2vec: Capturing network dynamics using dynamic graph representation learning}.
\newblock \bibinfo{journal}{\emph{Knowledge-Based Systems}}  \bibinfo{volume}{187} (\bibinfo{year}{2020}), \bibinfo{pages}{104816}.
\newblock


\bibitem[Hamilton et~al\mbox{.}(2017a)]%
        {hamilton2017inductive}
\bibfield{author}{\bibinfo{person}{Will Hamilton}, \bibinfo{person}{Zhitao Ying}, {and} \bibinfo{person}{Jure Leskovec}.} \bibinfo{year}{2017}\natexlab{a}.
\newblock \showarticletitle{Inductive representation learning on large graphs}.
\newblock \bibinfo{journal}{\emph{Advances in neural information processing systems}}  \bibinfo{volume}{30} (\bibinfo{year}{2017}).
\newblock


\bibitem[Hamilton(2020)]%
        {hamilton2020graph}
\bibfield{author}{\bibinfo{person}{William~L Hamilton}.} \bibinfo{year}{2020}\natexlab{}.
\newblock \bibinfo{booktitle}{\emph{Graph representation learning}}.
\newblock \bibinfo{publisher}{Morgan \& Claypool Publishers}.
\newblock


\bibitem[Hamilton et~al\mbox{.}(2017b)]%
        {Hamilton2017RepresentationLO}
\bibfield{author}{\bibinfo{person}{William~L. Hamilton}, \bibinfo{person}{Rex Ying}, {and} \bibinfo{person}{Jure Leskovec}.} \bibinfo{year}{2017}\natexlab{b}.
\newblock \showarticletitle{Representation Learning on Graphs: Methods and Applications}.
\newblock \bibinfo{journal}{\emph{IEEE Data Eng. Bull.}}  \bibinfo{volume}{40} (\bibinfo{year}{2017}), \bibinfo{pages}{52--74}.
\newblock
\urldef\tempurl%
\url{https://api.semanticscholar.org/CorpusID:3215337}
\showURL{%
\tempurl}


\bibitem[Han et~al\mbox{.}(2021)]%
        {Han2021TransformerIT}
\bibfield{author}{\bibinfo{person}{Kai Han}, \bibinfo{person}{An Xiao}, \bibinfo{person}{Enhua Wu}, \bibinfo{person}{Jianyuan Guo}, \bibinfo{person}{Chunjing Xu}, {and} \bibinfo{person}{Yunhe Wang}.} \bibinfo{year}{2021}\natexlab{}.
\newblock \showarticletitle{Transformer in Transformer}. In \bibinfo{booktitle}{\emph{Neural Information Processing Systems}}.
\newblock
\urldef\tempurl%
\url{https://api.semanticscholar.org/CorpusID:232076027}
\showURL{%
\tempurl}


\bibitem[Jia et~al\mbox{.}(2024)]%
        {Jia}
\bibfield{author}{\bibinfo{person}{Y. Jia}, \bibinfo{person}{C. Zhang}, {and} \bibinfo{person}{S. Vosoughi}.} \bibinfo{year}{2024}\natexlab{}.
\newblock \showarticletitle{Aligning Relational Learning with Lipschitz Fairness. International Conference on Learning Representations}. In \bibinfo{booktitle}{\emph{ICLR}}.
\newblock


\bibitem[Kazemi et~al\mbox{.}(2020)]%
        {kazemi2020representation}
\bibfield{author}{\bibinfo{person}{Seyed~Mehran Kazemi}, \bibinfo{person}{Rishab Goel}, \bibinfo{person}{Kshitij Jain}, \bibinfo{person}{Ivan Kobyzev}, \bibinfo{person}{Akshay Sethi}, \bibinfo{person}{Peter Forsyth}, {and} \bibinfo{person}{Pascal Poupart}.} \bibinfo{year}{2020}\natexlab{}.
\newblock \showarticletitle{Representation learning for dynamic graphs: A survey}.
\newblock \bibinfo{journal}{\emph{Journal of Machine Learning Research}} \bibinfo{volume}{21}, \bibinfo{number}{70} (\bibinfo{year}{2020}), \bibinfo{pages}{1--73}.
\newblock


\bibitem[Kipf and Welling(2016)]%
        {kipf2016semi}
\bibfield{author}{\bibinfo{person}{Thomas~N Kipf} {and} \bibinfo{person}{Max Welling}.} \bibinfo{year}{2016}\natexlab{}.
\newblock \showarticletitle{Semi-supervised classification with graph convolutional networks}.
\newblock \bibinfo{journal}{\emph{arXiv preprint arXiv:1609.02907}} (\bibinfo{year}{2016}).
\newblock


\bibitem[Kumar et~al\mbox{.}(2019)]%
        {kumar2019predicting}
\bibfield{author}{\bibinfo{person}{Srijan Kumar}, \bibinfo{person}{Xikun Zhang}, {and} \bibinfo{person}{Jure Leskovec}.} \bibinfo{year}{2019}\natexlab{}.
\newblock \showarticletitle{Predicting dynamic embedding trajectory in temporal interaction networks}. In \bibinfo{booktitle}{\emph{Proceedings of the 25th ACM SIGKDD international conference on knowledge discovery \& data mining}}. \bibinfo{pages}{1269--1278}.
\newblock


\bibitem[Li et~al\mbox{.}(2024)]%
        {li2024fast}
\bibfield{author}{\bibinfo{person}{Weikai Li}, \bibinfo{person}{Zhiping Xiao}, \bibinfo{person}{Xiao Luo}, {and} \bibinfo{person}{Yizhou Sun}.} \bibinfo{year}{2024}\natexlab{}.
\newblock \showarticletitle{Fast Inference of Removal-Based Node Influence}. In \bibinfo{booktitle}{\emph{Proceedings of the ACM Web Conference 2024}}. \bibinfo{pages}{422--433}.
\newblock


\bibitem[Li et~al\mbox{.}(2023)]%
        {li2023zebra}
\bibfield{author}{\bibinfo{person}{Yiming Li}, \bibinfo{person}{Yanyan Shen}, \bibinfo{person}{Lei Chen}, {and} \bibinfo{person}{Mingxuan Yuan}.} \bibinfo{year}{2023}\natexlab{}.
\newblock \showarticletitle{Zebra: When temporal graph neural networks meet temporal personalized PageRank}.
\newblock \bibinfo{journal}{\emph{Proceedings of the VLDB Endowment}} \bibinfo{volume}{16}, \bibinfo{number}{6} (\bibinfo{year}{2023}), \bibinfo{pages}{1332--1345}.
\newblock


\bibitem[Li et~al\mbox{.}(2021)]%
        {Li2021DynamicGL}
\bibfield{author}{\bibinfo{person}{Zhuoling Li}, \bibinfo{person}{Gaowei Zhang}, \bibinfo{person}{Lingyu Xu}, {and} \bibinfo{person}{Jie Yu}.} \bibinfo{year}{2021}\natexlab{}.
\newblock \showarticletitle{Dynamic Graph Learning-Neural Network for Multivariate Time Series Modeling}.
\newblock \bibinfo{journal}{\emph{ArXiv}}  \bibinfo{volume}{abs/2112.03273} (\bibinfo{year}{2021}).
\newblock
\urldef\tempurl%
\url{https://api.semanticscholar.org/CorpusID:244920721}
\showURL{%
\tempurl}


\bibitem[Manessi et~al\mbox{.}(2017)]%
        {Manessi2017DynamicGC}
\bibfield{author}{\bibinfo{person}{Franco Manessi}, \bibinfo{person}{Alessandro Rozza}, {and} \bibinfo{person}{Mario Manzo}.} \bibinfo{year}{2017}\natexlab{}.
\newblock \showarticletitle{Dynamic Graph Convolutional Networks}.
\newblock \bibinfo{journal}{\emph{Pattern Recognit.}}  \bibinfo{volume}{97} (\bibinfo{year}{2017}).
\newblock
\urldef\tempurl%
\url{https://api.semanticscholar.org/CorpusID:16745566}
\showURL{%
\tempurl}


\bibitem[Pareja et~al\mbox{.}(2020)]%
        {pareja2020evolvegcn}
\bibfield{author}{\bibinfo{person}{Aldo Pareja}, \bibinfo{person}{Giacomo Domeniconi}, \bibinfo{person}{Jie Chen}, \bibinfo{person}{Tengfei Ma}, \bibinfo{person}{Toyotaro Suzumura}, \bibinfo{person}{Hiroki Kanezashi}, \bibinfo{person}{Tim Kaler}, \bibinfo{person}{Tao Schardl}, {and} \bibinfo{person}{Charles Leiserson}.} \bibinfo{year}{2020}\natexlab{}.
\newblock \showarticletitle{Evolvegcn: Evolving graph convolutional networks for dynamic graphs}. In \bibinfo{booktitle}{\emph{Proceedings of the AAAI conference on artificial intelligence}}, Vol.~\bibinfo{volume}{34}. \bibinfo{pages}{5363--5370}.
\newblock


\bibitem[Pei et~al\mbox{.}(2020)]%
        {Pei2020GeomGCNGG}
\bibfield{author}{\bibinfo{person}{Hongbin Pei}, \bibinfo{person}{Bingzhen Wei}, \bibinfo{person}{Kevin Chen-Chuan Chang}, \bibinfo{person}{Yu Lei}, {and} \bibinfo{person}{Bo Yang}.} \bibinfo{year}{2020}\natexlab{}.
\newblock \showarticletitle{Geom-GCN: Geometric Graph Convolutional Networks}.
\newblock \bibinfo{journal}{\emph{ArXiv}}  \bibinfo{volume}{abs/2002.05287} (\bibinfo{year}{2020}).
\newblock
\urldef\tempurl%
\url{https://api.semanticscholar.org/CorpusID:210843644}
\showURL{%
\tempurl}


\bibitem[Qi et~al\mbox{.}(2023)]%
        {Qi2023UnderstandingOO}
\bibfield{author}{\bibinfo{person}{Xianbiao Qi}, \bibinfo{person}{Jianan Wang}, {and} \bibinfo{person}{Lei Zhang}.} \bibinfo{year}{2023}\natexlab{}.
\newblock \showarticletitle{Understanding Optimization of Deep Learning}.
\newblock \bibinfo{journal}{\emph{ArXiv}}  \bibinfo{volume}{abs/2306.09338} (\bibinfo{year}{2023}).
\newblock
\urldef\tempurl%
\url{https://api.semanticscholar.org/CorpusID:259171775}
\showURL{%
\tempurl}


\bibitem[Rossi et~al\mbox{.}(2020)]%
        {rossi2020temporal}
\bibfield{author}{\bibinfo{person}{Emanuele Rossi}, \bibinfo{person}{Ben Chamberlain}, \bibinfo{person}{Fabrizio Frasca}, \bibinfo{person}{Davide Eynard}, \bibinfo{person}{Federico Monti}, {and} \bibinfo{person}{Michael Bronstein}.} \bibinfo{year}{2020}\natexlab{}.
\newblock \showarticletitle{Temporal graph networks for deep learning on dynamic graphs}.
\newblock \bibinfo{journal}{\emph{arXiv preprint arXiv:2006.10637}} (\bibinfo{year}{2020}).
\newblock


\bibitem[Skarding et~al\mbox{.}(2020)]%
        {Skarding2020FoundationsAM}
\bibfield{author}{\bibinfo{person}{Joakim Skarding}, \bibinfo{person}{Bogdan Gabrys}, {and} \bibinfo{person}{Katarzyna Musial}.} \bibinfo{year}{2020}\natexlab{}.
\newblock \showarticletitle{Foundations and Modeling of Dynamic Networks Using Dynamic Graph Neural Networks: A Survey}.
\newblock \bibinfo{journal}{\emph{IEEE Access}}  \bibinfo{volume}{9} (\bibinfo{year}{2020}), \bibinfo{pages}{79143--79168}.
\newblock
\urldef\tempurl%
\url{https://api.semanticscholar.org/CorpusID:218665617}
\showURL{%
\tempurl}


\bibitem[Skarding et~al\mbox{.}(2021)]%
        {skarding2021foundations}
\bibfield{author}{\bibinfo{person}{Joakim Skarding}, \bibinfo{person}{Bogdan Gabrys}, {and} \bibinfo{person}{Katarzyna Musial}.} \bibinfo{year}{2021}\natexlab{}.
\newblock \showarticletitle{Foundations and modeling of dynamic networks using dynamic graph neural networks: A survey}.
\newblock \bibinfo{journal}{\emph{iEEE Access}}  \bibinfo{volume}{9} (\bibinfo{year}{2021}), \bibinfo{pages}{79143--79168}.
\newblock


\bibitem[Su et~al\mbox{.}(2024)]%
        {su2024pres}
\bibfield{author}{\bibinfo{person}{Junwei Su}, \bibinfo{person}{Difan Zou}, {and} \bibinfo{person}{Chuan Wu}.} \bibinfo{year}{2024}\natexlab{}.
\newblock \showarticletitle{PRES: Toward Scalable Memory-Based Dynamic Graph Neural Networks}.
\newblock \bibinfo{journal}{\emph{arXiv preprint arXiv:2402.04284}} (\bibinfo{year}{2024}).
\newblock


\bibitem[Szegedy et~al\mbox{.}(2014)]%
        {Szegedy}
\bibfield{author}{\bibinfo{person}{C. Szegedy}, \bibinfo{person}{W. Zaremba}, \bibinfo{person}{I. Sutskever}, \bibinfo{person}{J. Bruna}, \bibinfo{person}{D. Erhan}, \bibinfo{person}{I. Goodfellow}, {and} \bibinfo{person}{R. Fergus}.} \bibinfo{year}{2014}\natexlab{}.
\newblock \showarticletitle{Intriguing properties of neural networks}. In \bibinfo{booktitle}{\emph{ICLR}}.
\newblock


\bibitem[Tyagi and Sharma(2020)]%
        {tyagi2020taming}
\bibfield{author}{\bibinfo{person}{Sahil Tyagi} {and} \bibinfo{person}{Prateek Sharma}.} \bibinfo{year}{2020}\natexlab{}.
\newblock \showarticletitle{Taming resource heterogeneity in distributed ml training with dynamic batching}. In \bibinfo{booktitle}{\emph{2020 IEEE International Conference on Autonomic Computing and Self-Organizing Systems (ACSOS)}}. IEEE, \bibinfo{pages}{188--194}.
\newblock


\bibitem[Veli{\v{c}}kovi{\'c} et~al\mbox{.}(2017)]%
        {velivckovic2017graph}
\bibfield{author}{\bibinfo{person}{Petar Veli{\v{c}}kovi{\'c}}, \bibinfo{person}{Guillem Cucurull}, \bibinfo{person}{Arantxa Casanova}, \bibinfo{person}{Adriana Romero}, \bibinfo{person}{Pietro Lio}, {and} \bibinfo{person}{Yoshua Bengio}.} \bibinfo{year}{2017}\natexlab{}.
\newblock \showarticletitle{Graph attention networks}.
\newblock \bibinfo{journal}{\emph{arXiv preprint arXiv:1710.10903}} (\bibinfo{year}{2017}).
\newblock


\bibitem[Wang et~al\mbox{.}(2024)]%
        {wang2024learning}
\bibfield{author}{\bibinfo{person}{Xixi Wang}, \bibinfo{person}{Bo Jiang}, \bibinfo{person}{Xiao Wang}, {and} \bibinfo{person}{Bin Luo}.} \bibinfo{year}{2024}\natexlab{}.
\newblock \showarticletitle{Learning Dynamic Batch-Graph Representation for Deep Representation Learning}.
\newblock \bibinfo{journal}{\emph{International Journal of Computer Vision}} (\bibinfo{year}{2024}), \bibinfo{pages}{1--22}.
\newblock


\bibitem[Wang et~al\mbox{.}(2021)]%
        {wang2021apan}
\bibfield{author}{\bibinfo{person}{Xuhong Wang}, \bibinfo{person}{Ding Lyu}, \bibinfo{person}{Mengjian Li}, \bibinfo{person}{Yang Xia}, \bibinfo{person}{Qi Yang}, \bibinfo{person}{Xinwen Wang}, \bibinfo{person}{Xinguang Wang}, \bibinfo{person}{Ping Cui}, \bibinfo{person}{Yupu Yang}, \bibinfo{person}{Bowen Sun}, {et~al\mbox{.}}} \bibinfo{year}{2021}\natexlab{}.
\newblock \showarticletitle{Apan: Asynchronous propagation attention network for real-time temporal graph embedding}. In \bibinfo{booktitle}{\emph{Proceedings of the 2021 international conference on management of data}}. \bibinfo{pages}{2628--2638}.
\newblock


\bibitem[Wu et~al\mbox{.}(2020b)]%
        {Wu2020GraphNN}
\bibfield{author}{\bibinfo{person}{Shiwen Wu}, \bibinfo{person}{Fei Sun}, \bibinfo{person}{Fei Sun}, {and} \bibinfo{person}{Bin Cui}.} \bibinfo{year}{2020}\natexlab{b}.
\newblock \showarticletitle{Graph Neural Networks in Recommender Systems: A Survey}.
\newblock \bibinfo{journal}{\emph{Comput. Surveys}}  \bibinfo{volume}{55} (\bibinfo{year}{2020}), \bibinfo{pages}{1 -- 37}.
\newblock
\urldef\tempurl%
\url{https://api.semanticscholar.org/CorpusID:226246289}
\showURL{%
\tempurl}


\bibitem[Wu et~al\mbox{.}(2020a)]%
        {wu2020comprehensive}
\bibfield{author}{\bibinfo{person}{Zonghan Wu}, \bibinfo{person}{Shirui Pan}, \bibinfo{person}{Fengwen Chen}, \bibinfo{person}{Guodong Long}, \bibinfo{person}{Chengqi Zhang}, {and} \bibinfo{person}{S~Yu Philip}.} \bibinfo{year}{2020}\natexlab{a}.
\newblock \showarticletitle{A comprehensive survey on graph neural networks}.
\newblock \bibinfo{journal}{\emph{IEEE transactions on neural networks and learning systems}} \bibinfo{volume}{32}, \bibinfo{number}{1} (\bibinfo{year}{2020}), \bibinfo{pages}{4--24}.
\newblock


\bibitem[Xu et~al\mbox{.}(2020)]%
        {xu2020inductive}
\bibfield{author}{\bibinfo{person}{Da Xu}, \bibinfo{person}{Chuanwei Ruan}, \bibinfo{person}{Evren Korpeoglu}, \bibinfo{person}{Sushant Kumar}, {and} \bibinfo{person}{Kannan Achan}.} \bibinfo{year}{2020}\natexlab{}.
\newblock \showarticletitle{Inductive representation learning on temporal graphs}.
\newblock \bibinfo{journal}{\emph{arXiv preprint arXiv:2002.07962}} (\bibinfo{year}{2020}).
\newblock


\bibitem[Yang et~al\mbox{.}(2023)]%
        {Yang2023DynamicGR}
\bibfield{author}{\bibinfo{person}{Leshanshui Yang}, \bibinfo{person}{Cl{\'e}ment Chatelain}, {and} \bibinfo{person}{S{\'e}bastien Adam}.} \bibinfo{year}{2023}\natexlab{}.
\newblock \showarticletitle{Dynamic Graph Representation Learning With Neural Networks: A Survey}.
\newblock \bibinfo{journal}{\emph{IEEE Access}}  \bibinfo{volume}{12} (\bibinfo{year}{2023}), \bibinfo{pages}{43460--43484}.
\newblock
\urldef\tempurl%
\url{https://api.semanticscholar.org/CorpusID:258079246}
\showURL{%
\tempurl}


\bibitem[Ying et~al\mbox{.}(2018)]%
        {Ying2018GraphCN}
\bibfield{author}{\bibinfo{person}{Rex Ying}, \bibinfo{person}{Ruining He}, \bibinfo{person}{Kaifeng Chen}, \bibinfo{person}{Pong Eksombatchai}, \bibinfo{person}{William~L. Hamilton}, {and} \bibinfo{person}{Jure Leskovec}.} \bibinfo{year}{2018}\natexlab{}.
\newblock \showarticletitle{Graph Convolutional Neural Networks for Web-Scale Recommender Systems}.
\newblock \bibinfo{journal}{\emph{Proceedings of the 24th ACM SIGKDD International Conference on Knowledge Discovery \& Data Mining}} (\bibinfo{year}{2018}).
\newblock
\urldef\tempurl%
\url{https://api.semanticscholar.org/CorpusID:46949657}
\showURL{%
\tempurl}


\bibitem[Zhang et~al\mbox{.}(2020)]%
        {Zhang2020ResNeStSN}
\bibfield{author}{\bibinfo{person}{Hang Zhang}, \bibinfo{person}{Chongruo Wu}, \bibinfo{person}{Zhongyue Zhang}, \bibinfo{person}{Yi Zhu}, \bibinfo{person}{Zhi-Li Zhang}, \bibinfo{person}{Haibin Lin}, \bibinfo{person}{Yue Sun}, \bibinfo{person}{Tong He}, \bibinfo{person}{Jonas~W. Mueller}, \bibinfo{person}{R. Manmatha}, \bibinfo{person}{Mu Li}, {and} \bibinfo{person}{Alex Smola}.} \bibinfo{year}{2020}\natexlab{}.
\newblock \showarticletitle{ResNeSt: Split-Attention Networks}.
\newblock \bibinfo{journal}{\emph{2022 IEEE/CVF Conference on Computer Vision and Pattern Recognition Workshops (CVPRW)}} (\bibinfo{year}{2020}), \bibinfo{pages}{2735--2745}.
\newblock
\urldef\tempurl%
\url{https://api.semanticscholar.org/CorpusID:215828258}
\showURL{%
\tempurl}


\bibitem[Zhao et~al\mbox{.}(2021)]%
        {Zhao}
\bibfield{author}{\bibinfo{person}{X. Zhao}, \bibinfo{person}{Z. Zhang}, \bibinfo{person}{Z. Zhang}, \bibinfo{person}{L. Wu}, \bibinfo{person}{J. Jin}, \bibinfo{person}{Y. Zhou}, \bibinfo{person}{R. Jin}, \bibinfo{person}{D. Dou}, {and} \bibinfo{person}{D. Yan}.} \bibinfo{year}{2021}\natexlab{}.
\newblock \showarticletitle{Expressive 1-Lipschitz neural networks for robust multiple graph learning against adversarial attacks}. In \bibinfo{booktitle}{\emph{ICML}}. \bibinfo{pages}{12719–12735}.
\newblock


\bibitem[Zhou et~al\mbox{.}(2022)]%
        {zhou2022tgl}
\bibfield{author}{\bibinfo{person}{Hongkuan Zhou}, \bibinfo{person}{Da Zheng}, \bibinfo{person}{Israt Nisa}, \bibinfo{person}{Vasileios Ioannidis}, \bibinfo{person}{Xiang Song}, {and} \bibinfo{person}{George Karypis}.} \bibinfo{year}{2022}\natexlab{}.
\newblock \showarticletitle{Tgl: A general framework for temporal gnn training on billion-scale graphs}.
\newblock \bibinfo{journal}{\emph{arXiv preprint arXiv:2203.14883}} (\bibinfo{year}{2022}).
\newblock


\bibitem[Zhou et~al\mbox{.}(2020)]%
        {zhou2020graph}
\bibfield{author}{\bibinfo{person}{Jie Zhou}, \bibinfo{person}{Ganqu Cui}, \bibinfo{person}{Shengding Hu}, \bibinfo{person}{Zhengyan Zhang}, \bibinfo{person}{Cheng Yang}, \bibinfo{person}{Zhiyuan Liu}, \bibinfo{person}{Lifeng Wang}, \bibinfo{person}{Changcheng Li}, {and} \bibinfo{person}{Maosong Sun}.} \bibinfo{year}{2020}\natexlab{}.
\newblock \showarticletitle{Graph neural networks: A review of methods and applications}.
\newblock \bibinfo{journal}{\emph{AI open}}  \bibinfo{volume}{1} (\bibinfo{year}{2020}), \bibinfo{pages}{57--81}.
\newblock


\bibitem[Zhou et~al\mbox{.}(2018)]%
        {zhou2018dynamic}
\bibfield{author}{\bibinfo{person}{Lekui Zhou}, \bibinfo{person}{Yang Yang}, \bibinfo{person}{Xiang Ren}, \bibinfo{person}{Fei Wu}, {and} \bibinfo{person}{Yueting Zhuang}.} \bibinfo{year}{2018}\natexlab{}.
\newblock \showarticletitle{Dynamic network embedding by modeling triadic closure process}. In \bibinfo{booktitle}{\emph{Proceedings of the AAAI conference on artificial intelligence}}, Vol.~\bibinfo{volume}{32}.
\newblock


\bibitem[Zhuang et~al\mbox{.}(2023)]%
        {10.1145/3587268}
\bibfield{author}{\bibinfo{person}{Jiabo Zhuang}, \bibinfo{person}{Shunmei Meng}, \bibinfo{person}{Jing Zhang}, {and} \bibinfo{person}{Victor~S. Sheng}.} \bibinfo{year}{2023}\natexlab{}.
\newblock \showarticletitle{Contrastive Learning Based Graph Convolution Network for Social Recommendation}.
\newblock \bibinfo{journal}{\emph{ACM Trans. Knowl. Discov. Data}} \bibinfo{volume}{17}, \bibinfo{number}{8}, Article \bibinfo{articleno}{120} (\bibinfo{date}{June} \bibinfo{year}{2023}), \bibinfo{numpages}{21}~pages.
\newblock
\showISSN{1556-4681}
\href{https://doi.org/10.1145/3587268}{doi:\nolinkurl{10.1145/3587268}}


\end{thebibliography}

\appendix

\end{document}